\documentclass[twoside]{IEEEtran}
\usepackage{cite}
\usepackage{amsmath,amssymb,amsfonts}
\usepackage{algorithmic}
\usepackage{graphicx}
\usepackage{textcomp}
\usepackage{xcolor}

\usepackage{microtype}
\usepackage{graphicx}
\usepackage{subfigure}
\usepackage{booktabs} 
\usepackage{listings}

\usepackage{fancyhdr}

\usepackage{hyperref}

\def\BibTeX{{\rm B\kern-.05em{\sc i\kern-.025em b}\kern-.08em
    T\kern-.1667em\lower.7ex\hbox{E}\kern-.125emX}}
\begin{document}

\title{Flatland: a Lightweight First-Person 2-D Environment for Reinforcement Learning}

\author{\IEEEauthorblockN{Hugo Caselles-Dupr\'e\IEEEauthorrefmark{1}\IEEEauthorrefmark{2}, Louis Annabi\IEEEauthorrefmark{1}, Oksana Hagen\IEEEauthorrefmark{1}\IEEEauthorrefmark{3}, Michael Garcia-Ortiz\IEEEauthorrefmark{1}, David Filliat\IEEEauthorrefmark{2}} \\
\IEEEauthorblockA{\IEEEauthorrefmark{1} AI Lab (Softbank Robotics Europe)\\
\IEEEauthorrefmark{2} Flowers Laboratory (ENSTA ParisTech \& INRIA)\\
\IEEEauthorrefmark{3} Centre for Robotics and Neural Systems (Plymouth University)\\
Emails: caselles@ensta.fr, louis.annabi@gmail.com, oksana.hagen@softbankrobotics.com, \\ mgarciaortiz@softbankrobotics.com, david.filliat@ensta-paristech.fr\\}}

\maketitle
\thispagestyle{fancy}

 \newcommand{\david}[1]{\textcolor{teal}{D:#1}}




\begin{abstract}

\textit{Flatland} is a simple, lightweight environment for fast prototyping and testing of reinforcement learning agents. It is of lower complexity compared to similar 3D platforms (e.g. DeepMind Lab or VizDoom), but emulates physical properties of the real world, such as continuity, multi-modal partially-observable states with first-person view and coherent physics.
We propose to use it as an intermediary benchmark for problems related to Lifelong Learning. 
\textit{Flatland} is highly customizable and offers a wide range of task difficulty to extensively evaluate the properties of artificial agents.
We experiment with three reinforcement learning baseline agents and show that they can rapidly solve a navigation task in \textit{Flatland}. A video of an agent acting in \textit{Flatland} is available here: \url{https://youtu.be/I5y6Y2ZypdA}.

\end{abstract}

\section{Introduction}

 A key goal of artificial intelligence research is to design agents capable of Lifelong Learning, which is the continued learning of tasks, from one or more domains, over the course of a lifetime of an artificial agent. Emulation of a real-life sensorimotor experience is an important aspect of Lifelong Learning. Such emulation enables building agents, that are capable of acquiring common sense knowledge in form of learning naive physics \cite{DBLP:journals/corr/AgrawalNAML16} or by building models of sensorimotor interactions with the environment \cite{merlin, Ha2018WorldModels}.


With this goal in mind, several complex 3-D partially-observable environments have been developed, such as DeepMind Lab \cite{beattie2016deepmind}, VizDoom \cite{kempka2016vizdoom} or Malmo \cite{johnson2016malmo}. They are used for testing Reinforcement Learning (RL) agents on tasks and scenarios that require advanced capabilities in terms of perception, planning, representation of space for navigation-related tasks, which paves the way for building agents capable of behaving in a real-life scenario. These environments are suitable for Lifelong Learning research since they emulate key features of real-life tasks such as first-person, partial-observability and coherent physics. However we argue that these benchmarks have limitations (see Sec.\ref{sec:related}). The high richness of sensors and environment in current 3-D immersive simulations does not allow for fast experiments and prototyping. 

The environments used for prototyping, e.g grid-worlds, do not have the features required for building agents capable of behaving in real-life scenarios, such as coherent physics or continuous state and action space. There is a need for a middle ground benchmark for testing artificial agents, which preserves the complexity of the tasks, while reducing the complexity of the sensory space.


We propose \textit{Flatland}, a 2-D first-person view environment where an agent can move and interact with different elements of the environment, constrained by simple physical laws (see Sec.\ref{sec:flatland}). All the obstacles, objects and agents may have different physical properties and visual appearances. With a two dimensional world, we reduce the complexity of the environment while preserving key features of the physical world, such as first-person, partial-observability and coherent physics.
Our environment allows to conduct fast RL experiments, as we demonstrate on a navigation task (see Sec.\ref{sec:experiments}) . We show that three baseline RL algorithms solve this task two orders of magnitude faster than a similar experiment \cite{dosovitskiy2016learning} in VizDoom. Faster experiments allow researchers to perform extensive testing which, as pointed out in \cite{henderson2017deep}, is crucial for reproducibility and establishing the significance of the results. 

We plan on releasing \textit{Flatland} as open-source software. The software is easily expandable and is compatible with OpenAI Gym RL API \cite{1606.01540}. Future versions of the simulator will be guided by feedback and requests from the community.

\section{Related work}\label{sec:related}

To experiment and develop new artificial agents, it is common to use grid-world simulations \cite{sutton1998reinforcement, barreto2017successor, teh2017distral} as they are easy to manipulate, fast to train, and easily comprehensible for an external observer.  However, grid-world environments often assume a perfect knowledge of the agent's state. This simplified situation does not tie in with real-life scenarios and makes it impossible to tackle issues related to the grounding of perception or the emergence of common sense knowledge \cite{lake2017building}.

The Atari suite \cite{bellemare2013arcade} provides environments where the agent perceives its state through 2D images, and has a small set of discrete actions that it can perform to play a game. These environments are more complex than grid-worlds and allow to develop new algorithms for control and planning. Still, they are fully observable and thus do not correspond to real-life scenarios.

Recent simulators allow to generate environments closer to real-life experience. It allows to test approach without having to deal with the problems of robots in the real world: hardware failure, cost of experiments, time of experiments, scalability. For example, RL benchmarks such as MuJoCo \cite{todorov2012mujoco} focus on the building of control policies for robots. In these simulators, the agent performs complex continuous actions in a realistic physical world in order to attain a certain position or a certain speed. These benchmarks helped develop new control algorithms for, e.g., bipedal walking.

To simulate real-life scenarios, the appropriate simulators are partially-observable and first-person 3-D environments (such as DeepMind Lab or VizDoom). They allow to tackle the challenges related to navigation and planning with a first-person sensory input. In these environment, richness of sensors and actions (2D images, continuous actions), as well as task complexity (need for memory and long-term prediction), make the computation required to train algorithms prohibitive for most research teams. The computational cost of the experiments on realistic environments does not facilitate principled statistical analysis of the proposed approaches. The same argument could be made for task complexity, as it becomes impossible to test a large variety of tasks using these environments.

We believe that a first-person partially observable environment with an intermediary level of sensory richness and tasks of increasing complexity would help the community to test and evaluate new approaches that are ultimately designed to be applied for acting in real-life scenarios.

\section{\textit{Flatland}: a 2-D simulator}\label{sec:flatland}

\begin{figure}[ht]
\vskip 0.2in
\begin{center}
\centerline{\includegraphics[scale=0.5]{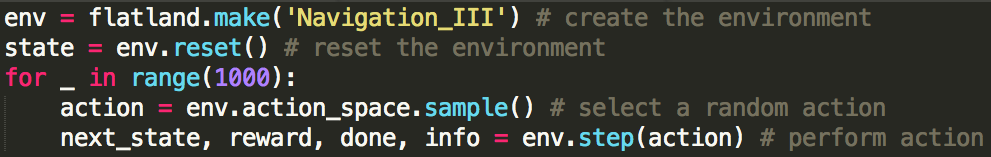}}
\caption{\textit{Flatland}'s python API. It is identical to the OpenAI Gym \cite{1606.01540} API such that researchers used to the latter can use the former easily.}
\label{flatland_api}
\end{center}
\vskip -0.2in
\end{figure}

\textit{Flatland} is a first-person 2-D game platform designed for research and development of autonomous artificial agents. \textit{Flatland} can be used to study how artificial agents may learn tasks in partially or fully observable diverse worlds. The game engine is based on Pymunk and Pygame Python libraries and inspired from a blog post from Matt Harvey \cite{rlcar}. The tasks are inspired from the DeepMind Lab environment \cite{beattie2016deepmind}. 

We design \textit{Flatland} for fast prototyping and testing of ideas related to Lifelong Learning such as acquiring common sense knowledge, which is  built upon learning about regularities in the sensorimotor information available to the agent \cite{Friston2009TheFPKarl}, and constructing sensory representations as well as a model of the physics of the environment. RL provides a convenient framework for evaluation of autonomous open-ended learning, and can be used to quantify and compare different artificial agents. Therefore, \textit{Flatland} is designed to be used in the context of RL algorithms.

The simulator creates a rectangular environment with several physical objects with different physical properties and textures. Environments are composed of rooms and corridors which are delimited by walls or doors and may contain obstacles and items that can be picked up to obtain (positive or negative) rewards. All these variables can be specified by the user so that tasks of a wide range of difficulty can be considered. Environments are generated based on a human-readable configuration file.

Regarding state spaces, the agent perceives through one or several first-person sensors which provide, for example, information about the color or distance of elements of its environment (e.g. depth and RGB sensors), or top-down view of an area around the agent. In the experiments presented hereafter, the agent has one sensory input corresponding to colors with an intensity depending on distance (see Fig.~\ref{inputs}). 

Concerning action spaces, the agents can act discretely (go forward, turn left, turn right for navigation for instance) or continuously (one continuous dimension for each movement). Additionally, the simulator allows to easily attach body parts (arms and hooks) to the agent if we want to implement more complex tasks. Games can be composed of more than one agent, which can be interesting to study the emergence of collaborative or social behaviours in multi-agent systems.

We ensure compatibility with OpenAI Gym RL API by defining the same API, as illustrated in a minimal example in Fig.~\ref{flatland_api}. 



\section{Experiments}\label{sec:experiments}

We conduct preliminary experiments to evaluate the potential of \textit{Flatland} as a fast prototyping platform for learning agents.

\subsection{A test navigation task}\label{task}

\begin{figure}[h]
\vskip 0.2in
\begin{center}
\centerline{\includegraphics[scale=0.25]{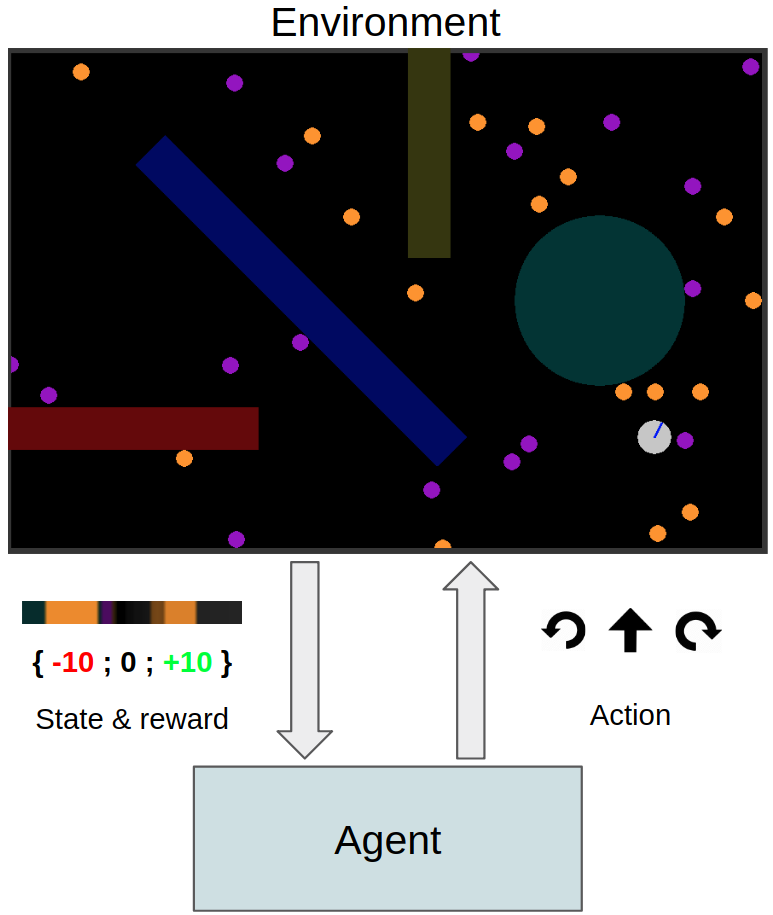}}
\caption{Experimental setup. The agent's task is to collect orange ``fruits" ($+10$ reward) while avoiding purple ``poisons" ($-10$ reward). The agent's inputs are 1-D images, which are extended to 2-D images here for visualization purposes.}
\label{inputs}
\end{center}
\vskip -0.2in
\end{figure}

\begin{figure*}
\includegraphics[width=\textwidth]{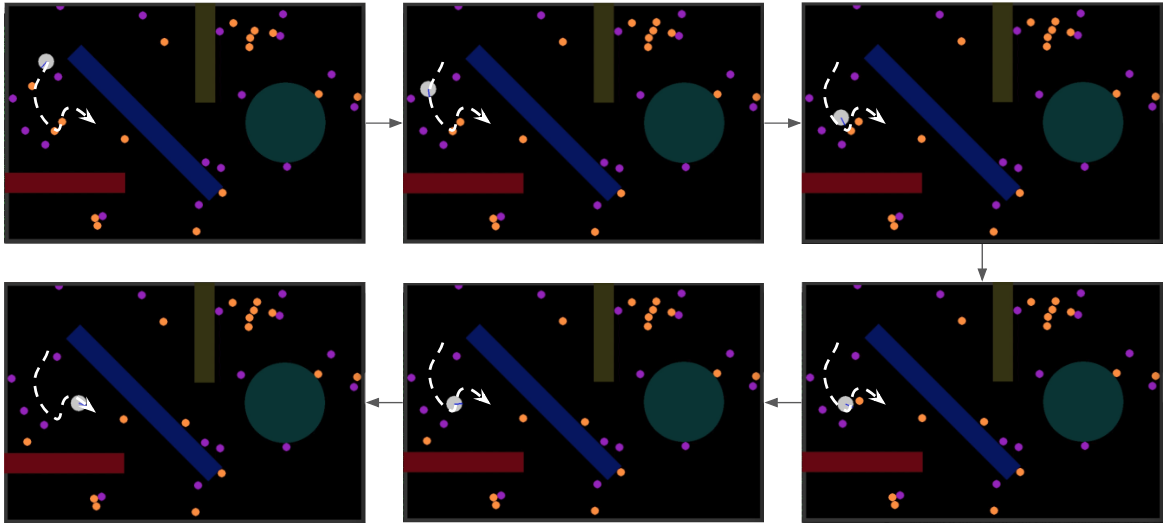}
\caption{Evidence of non-trivial behavior of DQN. The agent carefully navigates in the environment in order to collect ``fruits" while avoiding ``poisons". Best viewed in color. 
}
\label{behaviour}
\end{figure*}

For this task we have defined an environment that consists of a room with 4 obstacles of different shapes and colors, and two types of objects that can be picked up. The agent can collect orange ``fruits" ($+10$ reward) or purple ``poisons" ($-10$ reward) by touching them. No other reward signal is provided. The goal of the agent is to find a strategy to get a maximum reward in a fixed amount of timesteps ($500$). It receives as input a 1-D image corresponding to what the agent sees in front of it, which is processed using 1-D convolutional layers. This task is similar to the one proposed for the evaluation in \cite{dosovitskiy2016learning} and \cite{mnih2016asynchronous}.  
The agent's action space is discrete and composed of 3 actions: \textit{move forward}, \textit{rotate left} and \textit{rotate right}.
The task, inputs and actions are illustrated in Figure \ref{inputs}.\\

\subsection{Baseline models}

We propose to test our simulator by solving the task described in Sec.\ref{task} using 3 baselines that were previously evaluated on a first-person view navigation tasks: \textbf{Deep Q-Network} (DQN) \cite{mnih2015human}, \textbf{Asynchronous Advantage Actor-Critic} (A3C) \cite{mnih2016asynchronous}, two model-free RL baselines, and \textbf{Direct Future Prediction} (DFP) \cite{dosovitskiy2016learning}, an algorithm that learns to act by predicting features of the environment. The goal of this experiment is not to compare merits and weaknesses of the three evaluated methods. We rather show that three RL baselines perform well on a navigation task in \textit{Flatland}, and do converge faster comparing to complex 3D environments.


DQN \cite{mnih2015human} combines Q-learning \cite{Watkins:1989} with experience replay \cite{lin1992self}  and a deep neural network. For a given state $s$, DQN outputs a vector of action values $Q(s, \cdot , \theta)$, where $\theta$ are the parameters of the network. Policies are derived using an $\epsilon$-greedy approach w.r.t to $Q$. DQN was tested on a set of Atari 2600 games \cite{bellemare2013arcade}, reaching human-level performance on many games. It has thus become a commonly used baseline for RL problems. 


In A3C \cite{mnih2016asynchronous}  many instances of the agent interact in parallel with many instances of the environment, which both accelerates and stabilizes learning.  The A3C algorithm constructs an approximation to both the policy $\pi(a|s, \theta)$ and the value function $V(s, \theta_v)$ using parameters of the actor $\theta$ and parameters of the critic $\theta_v$. Both policy and value are adjusted towards an n-step lookahead value using an entropy regularization penalty. 


In DFP \cite{dosovitskiy2016learning}, the reinforcement learning problem is reformulated as a supervised learning  problem. The agent predicts high-level features at multiple timescales (here: score, number of ``fruits" picked up and number of ``poisons" picked up), which are combined linearly to form an objective function. Then, policies are derived using actions that maximize this objective. This algorithm won, by a large margin, the VizDoom 2016 competition.

\subsection{Experimental setup}

We use available architectures for each method, and replace 2-D convolutions by 1-D convolutions. In Appendix \ref{sec:appendixa}, we provide all hyperparameters and references to used implementations. Each method is trained for $500$ episodes ($250$k timesteps) using the Adam optimizer \cite{kingma2014adam}. Reported results and figures show results averaged over $30$ seeds, following recommendations in \cite{henderson2017deep} of evaluating on more than a large enough sample of random seeds to ensure significant results. 


\section{Results}

\begin{figure}[ht!]
\vskip 0.2in
\begin{center}
\centerline{\includegraphics[scale=0.6]{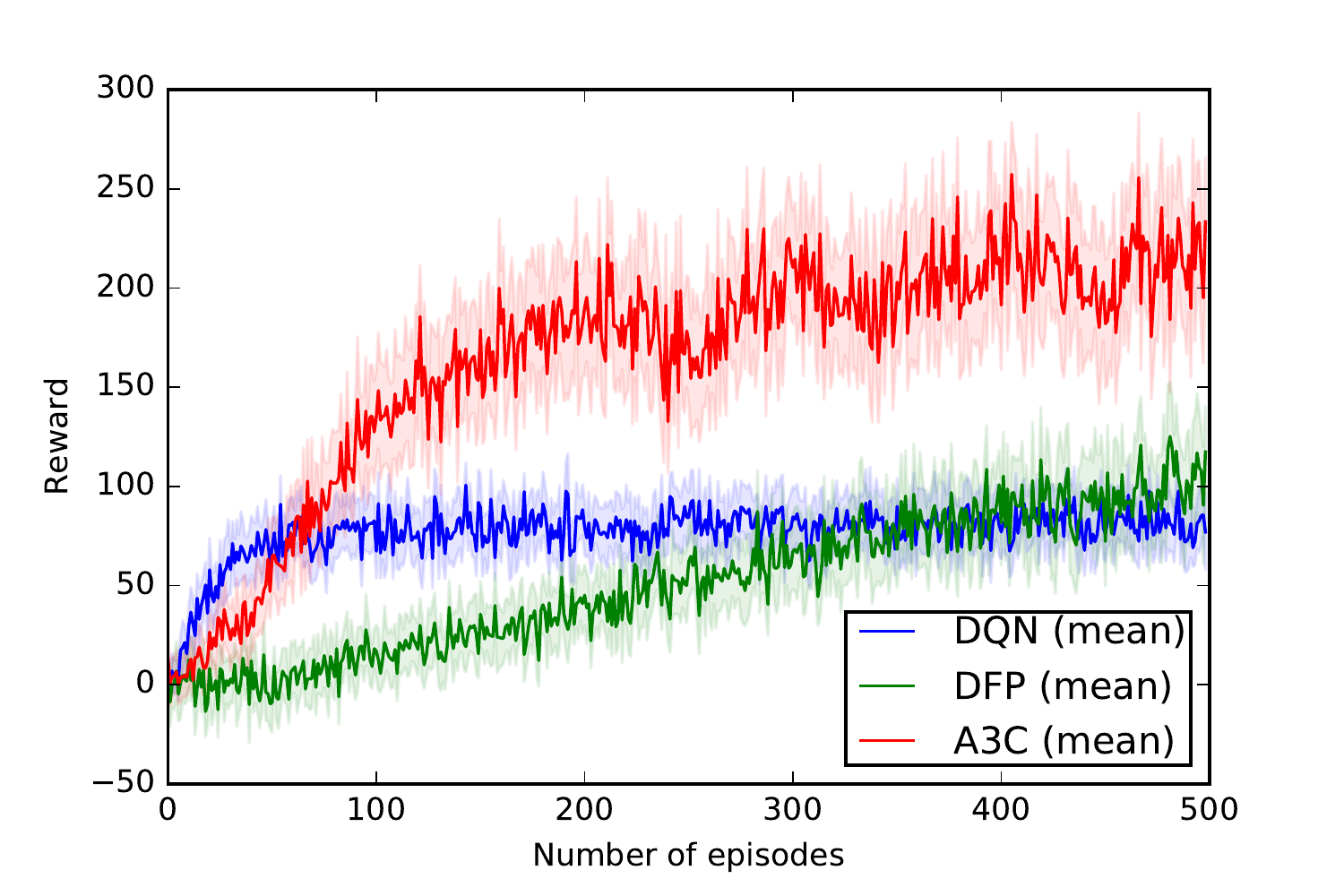}}
\caption{Training curves for DQN, A3C and DFP. Means, and 95\% confidence intervals are computed using $30$ independent runs.}
\label{results}
\end{center}
\vskip -0.2in
\end{figure}

Training results are reported in Figure \ref{results}. We report the average reward as well as $95\%$ confidence interval computed with a standard t-test for the mean with a sample size of $30$, as recommended in \cite{hogg2009probability}.

All three methods rapidly converge to a policy that successfully navigates in the environment to pick up ``fruits" while avoiding ``poisons". Agents learn to navigate efficiently. This is illustrated in Fig. \ref{behaviour} and in the video included in Appendix \ref{sec:appendixb}. The policies are completely reactive since none of the methods incorporates memory. The best scoring agents achieved scores close to that obtained by humans after playing several games ($\approx300$ reward).

In a similar navigation task in the experiments of \cite{dosovitskiy2016learning}, convergence is obtained after roughly $20$M timesteps for DQN, DFP and A3C. Here, convergence is reached approximately two order of magnitudes faster with roughly $200$k timesteps. Since the task complexity is similar, these results confirm that faster convergence can be obtained using lower-dimensional sensory stream. This faster convergence is two fold. First the 1-D images sensors contain less rich information than 2-D images so the training is faster in terms of training steps. Second it allows for a reduced number of parameters for the neural network (1-D instead of 2-D convolutional layers), hence a faster optimization of the neural network in terms of wall-clock time. Overall this allows for experiments at a reduced computational cost. For instance, training DQN took less than 1 hour on a Intel i7-7700K CPU.  


\section{Conclusion}

We propose \textit{Flatland}, a lightweight, 2-D, partially-observable, highly flexible environment for testing and evaluation of reinforcement learning agents, in the context of problems related to Lifelong Learning. This simulation preserves main characteristics of the tasks that agents face in real life environments, such as partial observability, coherent physics and a first-person view. However, contrary to commonly used complex 3-D benchmarks with the same properties, our environment has a much lower dimensional sensory stream. This enables fast prototyping and extensive experiments at low cost, as we show by experimenting with three RL baselines in a navigation task. On this task, convergence is achieved two order of magnitude faster, in terms of number of training steps, compared to a similar experiment on VizDoom. \textit{Flatland} is designed as a framework for the evaluation of the performance of RL agents, and features OpenAI Gym API.

The version of \textit{Flatland} presented in this paper is preliminary. Future work on \textit{Flatland} includes adding head and arms to agents, so that they are able to manipulate the environment by moving objects. This would allow us to define more complex tasks, which could potentially require reasoning and memory to be solved. Another direction of future work includes adding more diverse sensors (e.g touch, smell), more actions (e.g. grab objects, open doors). This would allow us to implement multi-modal learning, which is a promising direction in artificial intelligence research. The space of the possible tasks could be expanded by adding more features, such as defining tasks with delayed rewards. Finally, we also intend to enable the ability to add multiple agents to the simulation, so that this environment could be used to study multi-agent interactions. 

Most importantly, this benchmark should evolve according to the community's needs in terms of environment's nature and characteristics.


\newpage

\appendix
\section{Implementations details}\label{sec:appendixa}

We present implementation details for each of the three RL baselines that we experiment with (see Sec. \ref{sec:experiments} of main paper).
\subsection{Deep Q-Network}

\begin{itemize}
\item Implementation: Keras-rl (\url{https://github.com/keras-rl/keras-rl})
\item Normalization of inputs
\item Adam: $learning\_rate = 0.001$, $\beta _1=0.9$, $\beta _2=0.999$
\item Policy: Boltzmann policy (softmax) with temperature 1.
\item 1500 timesteps warmup
\item Soft updates for Q-learning parameter: 0.01
\item Replay buffer size: 500000
\item Architecture: Input (shape = (64,3)) - Convolution 1-D (filters: 32, kernel size: 8, 1) - Convolution 1-D (48,4,1) - Convolution 1-D (64,3,1) - Max Pooling 1-D - Dense layer (3) - Output (shape = 3)
\item Discount factor of $\gamma = 0.99$
\end{itemize}

\subsection{Asynchronous Advantage Actor-Critic}

\begin{itemize}
\item Implementation: \url{https://github.com/openai/universe-starter-agent}
\item Adam: $learning\_rate = 0.00001$, $\beta _1=0.9$, $\beta _2=0.999$
\item 5 actor-learner threads, all methods performed updates after every 20 actions ($t_{max} = 20$ and $I_{update} = 20$).
\item Entropy regularization with a weight $\beta = 0.01$
\item Discount of $\gamma = 0.99$
\item No action repeat: execute action on every frame (action repeat = 1)
\item Architecture: Convolution 1-D (filters: 32, kernel size: 8, 1) - Convolution 1-D (48,4,1) - Convolution 1-D (64,3,1), we have substituted the LSTM with a fully connected layer of dimension 128 to make the policy reactive.
\end{itemize}

\subsection{Direct Future Prediction}

\begin{itemize}
\item Implementation: \url{https://github.com/flyyufelix/Direct-Future-Prediction-Keras}
\item Adam: $learning\_rate = 0.00001$, $\beta _1=0.9$, $\beta _2=0.999$
\item Measurements used: score, number of fruits picked up, number of poisons picked up
\item Goal: [1,1,-1]
\item Normalization of inputs and measurements
\item 1000 timesteps warmup
\item Training interval: 3 timesteps
\item Policy: $\epsilon$-greedy with respect to the defined goal
\item Linear annealing of epsilon from 1 to 0.0001: $\epsilon = (\epsilon_{final} - \epsilon_{initial})/300000$
\item Replay buffer size: 20000
\item Architecture: We only modify the convolutional part with: Convolution 1-D (filters: 32, kernel size: 8, 1) - Convolution 1-D (48,4,1) - Convolution 1-D (64,3,1) - Max Pooling 1-D, the rest is unchanged.
\item Discount factor of $\gamma = 0.99$
\end{itemize}

\section{Supplementary video}\label{sec:appendixb}

We provide a supplementary video to illustrate the behaviour of a Deep Q-Network agent after learning in the experiments. The video can be found here: \url{https://youtu.be/I5y6Y2ZypdA}.

\end{document}